# Unsupervised Stemming based Language Model for Telugu Broadcast News Transcription

Pala Mythilisharan, Parayitam Laxminarayana and Appala Venkataramana
Osmania University
Hyderabad - India

Abstract :- In Indian Languages , native speakers are able to understand new words formed by either combining or modifying root words with tense, number and / or gender. Due to data insufficiency, Automatic Speech Recognition system (ASR) may not accommodate all the words in the language model irrespective of the size of the text corpus. It also becomes computationally challenging if the volume of the data increases exponentially due to morphological changes to the root word.

In this paper a new unsupervised method is proposed for one Indian language, Telugu, based on the unsupervised method for Hindi, in order to generate the Out of Vocabulary (OOV) words in the language model. By using techniques like smoothing and interpolation of pre-processed data with supervised and unsupervised stemming, different issues in language model for Telugu has been addressed. We observe that the smoothing techniques Witten-Bell and Kneser-Ney perform well when compared to other techniques on pre-processed data from supervised learning. The ASRs accuracy is improved by 0.76% and 0.94% with supervised and unsupervised stemming respectively.

I. INTRODUCTION

Telugu is one of four modern literary languages belonging to the Dravidian family. It is also one of the six classical languages of India. With native speakers of 81 million, it stands as fourth most widely spoken language in the sub-continent. A comprehensive ASR for Telugu language has not been made available due to lack of standard publicly accessible annotated speech corpus[1].

ASR leverages acoustic model, language model and lexicons for recognition accuracy. Language model gives the distribution of probabilities on sequence of words, calculated using the available training text corpus. Test speech may contain few new words that may not have been acquainted in the training corpus. These new words cannot be recognized by the decoder. New words appearing in test speech are called as Out of Vocabulary (OOV). Complexity of OOV is more in Indian Languages than European Languages due to hybridization of multiple words transforming to one complex word in verbal communication.

OOV is a common phenomenon in the Large Vocabulary Continuous Speech Recognition (LVCSR) systems. The methods used in addressing the OOV for English are not suitable for orthographies of Indian languages. Pertinent techniques are required to solve the problem of OOVs for Indian Languages [2]. Since most Indian languages have agglutinative morphologies and are associated with phonemic orthography (akshara based orthographies) most words are inflected forms of their root counterparts. Recognition accuracy of ASR may be improvised with generation of OOVs for LM, using inflections or morphological modifications of existing root words. If newly generated words are included in the training text corpus to obtain Language Models, size of the vocabulary increases considerably. Therefore, it is proposed to include the root word and its prefixes or suffixes, as words in the training text corpus.

Same language models may be used for decoding speech into text. Wherein, the recognized text which appears with root word and prefix or suffix, is combined to form a meaningful new word. The performance of ASR is evaluated against supervised and unsupervised methods to generate new words with inflections or morphological modifications as reported in the literature[3].

The other issue to be addressed in the language model is regarding unseen sequence of words in the test speech. Language model is used in the ASR decoder, to constrain the search in various ways by giving favourable bias to the sequence of more probable words. It is a common phenomena to have unseen sequence of words while performing recognition. The performance of the recognizer will be degraded if unseen sequence of words appears in a test speech. In the LVCSR scenario, as per Zipf's laws [4]– irrespective of the size of text corpus used for training, it is not feasible to have all possible sequences of words. To reduce the effect of unseen sequence of words, various smoothing and discounting techniques can be employed for obtaining the language models. Well established techniques like Good-Turing, linear discounting, absolute discounting, Witten-Bell, and Kneser-Ney along with its modified version play a significant role in reducing the effect of unseen sequence of words. However, these methods are verified only for English and other European languages[5].

In this work, we build and evaluate existing LMs for Telugu language using supervised and unsupervised stemming. The Second section deals with building of annotated speech corpus, development of acoustic and language models for Telugu ASR. Third section deals with the analysis of performance of ASR accuracy with different smoothing and discounting techniques. Fourth section explores the details about supervised and unsupervised stemming techniques used for pre-processing the text corpus for creating language models and test their performance.

II. BUILDING ANNOTATED SPEECH, TEXT CORPUS AND ACOUSTIC AND LANGUAGE MODELS FOR ASR USED IN TELUGU BROADCAST NEWS

To evaluate the performance of a LVCSR system, large amount of annotated speech database under different environmental conditions and large text corpus are required for acoustic modelling and Language Model.

*A.    Speech Corpus*

As the application is subjected for transcribing Telugu broadcast news, sixty five hours of video data broadcasted by the Telugu TV channels is sourced from the YouTube system[[6]. Then the audio and video data is separated using open source tools [7]. The data consists of speech by news readers, anchors, expert analysts invited for discussion for news analysis, live news reports and reporter's interaction with political leaders, public etc. The speech data also consists of public addresses and snippets of discussions in legislative assembly, reported in the news bulletin. The speech data explained above comprises of voices of 298 female and 327 male speakers. This speech data after discarding the non-speech sounds from the audio stream is transcribed using the transliteration tool BARAHA [8]. The speech segments with cross talk in the assembly or TV discussions or in the reporting news from public places is also excluded from the training data.

Lexicon is a representation of each unique word in the text corpus by a sequence of its representative phonemes. In this experiment, unique words are generated from the transcribed text of speech data and text corpora that are collected for language modelling. Phonemes are acoustic units that are to be modelled. So, it is ensured that the training data covers all the phonemes for adequate number of times by choosing special speech files covering less frequently appearing phonemes. 60 phonemes are considered. They are listed in TABLE 1. The Indian Language Speech Label (ILSL12) set [9] with 47 items for Telugu is used for the development of Telugu phoneme set and lexicon. In addition to ILSL12 set, thirteen geminates or double consonants in Telugu script are added as additional phonemes, as they do not follow the same pattern in pronunciation[10][11].

TABLE 1 TABLE OF TELUGU PHONEMES

| Roman representation | Telugu font | Roman representation | Telugu font | Roman representation | Telugu font |
|---|---|---|---|---|---|
| a | అ | K | క | m | మ |
| aa | ఆ | kh | ఖ | y | య |
| i | ఇ | g | గ | r | ర |
| ii | ఈ | gh | ఘ | l | ల |
| u | ఉ | c | చ | lx | ళ |
| uu | ఊ | ch | ఛ | w | వ |
| rq | ఋ | j | జ | sh | శ |
| rrq | ౠ | jh | ఝ | sx | ష |
| e | ఎ | nj | ఞ | h | హ |
| ei | ఏ | tx | ట | rx | ఱ |

| ai | ఐ | txh | థ | d | ద |
| o | ఒ | dx | డ | dh | ధ |
| oo | ఓ | dxh | ఢ | p | ప |
| au | ఔ | nx | ణ | ph | ఫ |
| mn | O | n | న | b | బ |
|    |   |    |   | bh | భ |
| geminate consonant phonemes | | | | | |
| kk | క్క | cc | చ్చ | txtx | ట్ట |
| gg | గ్గ | jj | జ్జ | dxdx | డ్డ |
| tt | త్త | pp | ప్ప | ss | స్స |
| dd | ద్ద | bb | బ్బ | nn | న్న |
| mm | మ్మ | ww | వ్వ |   |   |

### B. Acoustic Models

Acoustic models can be generated using Hidden Markov Models (HMM) or Subspace Gaussian Mixture Models (SGMM) or Deep Neural Networks (DNNs)[12][13][14]. Here, HMM and SGMM based acoustic models are considered for carrying out proposed study of performance of ASR accuracy for different smoothing techniques used for language model.

HMM and SGMM are also generated using the Kaldi toolkit[15] for ASR. Standard Mel-frequency cepstral coefficients (MFCCs) with delta and double delta coefficients are extracted and Linear Discriminant Analysis (LDA) and Maximum Likelihood Linear Transform (MLLT) feature-space transformations were applied on these MFCC features for feature-space reduction. Feature-space Maximum Likelihood Linear Regression (f-MLLR) is used for building Gaussian Mixture Models (GMM) for tied state triphones. These GMMs were subsequently used to build SGMMs again using Kaldi tool kit[13].

### III. PERFORMANCE OF ASR ACCURACY WITH DIFFERENT SMOOTHING TECHNIQUES FOR LANGUAGE MODELS

### A. Introduction to smoothing and discounting

Data sparsity is a common problem for many applications. Even though large amount of text data is available for language modeling, it only develops higher order N-gram models, which reduces sparsity to a certain extent but does not remove completely. So, the probability of N-gram sequences without appearing in the training text corpora will be assigned a zero in the language model. With higher order N-gram models, sparsity also increases. Here, the data sparsity refers to the N-grams which occur in test data but not in training data. While there are many techniques to address the sparse data, of them smoothing technique is more popular for addressing sparse data issues in the language modeling[16]. The structure of language model is unchanged but the method used to estimate the probabilities of the model is modified by smoothing. Thus, irrespective of the size of data, smoothing is required to handle the sparsity and improve the performance of language modeling and the performance of ASR system[17]. Smoothing techniques adjusts low probabilities such as zero probabilities upward and high probabilities downward. The smoothing techniques adjust the probabilities, such that the estimate of probabilities of the sequences should not deviate much from the true or original probabilities. Smoothing methods prevent not only zero probabilities, but also attempt to improve the accuracy of the model[18].

*B. Different smoothing techniques*

Many variations are reported in the literature for smoothing; Jelinek and Mercer [19]; Katz [20]; Bell et al.[21]  Ney et al. [16], and Kneser and Ney[22] [17]. Comparison of different smoothing techniques for different applications and different languages are reported in the literature[23][22]. However, published accounts of comparison with different smoothing techniques for Indian languages especially for Telugu is not found in the literature[18]. Therefore, there is a need to study and build efficient language models for for Telugu, with proper smoothing techniques. So in this project different smoothing techniques are evaluated with the transcription of Telugu TV news using ASR.

To generate the LMs for Telugu, transcribed text of Training and testing speech corpus and another 1,60,271 sentences consisting 19,41,832 words of Telugu text was collected from different online news websites. This text corpus is used for evaluation of smoothing models for language modeling. The number of unique words covered in the text data are 2,37,994.

The advanced smoothing techniques will assign nonzero probability for never seen things (sequence of words or individual words) based on probability of things appeared at least once in the training data. In this paper, popular smoothing techniques like, Linear discounting, Good-Turing, Absolute discounting, Witten-Bell, Kneser Ney discounting are considered for evaluation [24][25][26][22].

The cmuclmtk-0.7   [27] is used to generate the Linear discounting, Good-Turing, Absolute discounting and Witten-Bell. The cmuclmtk-0.7 which initially working for 65,535 words, is modified to work for 2,50,000 words. Using SRILM tool[28], we have generated Language model for   Kneser Ney smoothing. The performance of different smoothing techniques is given in **Error! Reference source not found.**.Witten-Bell smoothing performs better among all tested smoothing techniques. So Witten-Bell smoothing was selected for further analysis.

TABLE 2 WER with modifying the LM using language structure Total Data considered for testing is 2-hour 30 min as HMM as acoustic model

| S.No | Type of smoothing | Correct | Errors | | | | WER |
|------|-------------------|---------|--------|--|--|--|-----|
|      |                   |         | Insertions | Deletions | Substitutions | TOTAL |     |

| 1 | Good –Turing | 5091 | 68 | 534 | 1189 | 1791 | 26.28 |
| 2 | Linear discounting | 5084 | 61 | 555 | 1175 | 1791 | 26.28 |
| 3 | Absolute discounting | 5101 | 72 | 523 | 1190 | 1785 | 26.20 |
| 4 | Witten-Bell | 5376 | 73 | 437 | 1001 | 1511 | 22.17 |
| 5 | Kneser Ney | 5151 | 66 | 505 | 1158 | 1729 | 24.37 |

The most common metric for evaluating a language model is the probability that the model assigns to test data that is termed, perplexity [29]. Best language model obtained from the training data, is expected to give higher probabilities to the events occurred in the test data.

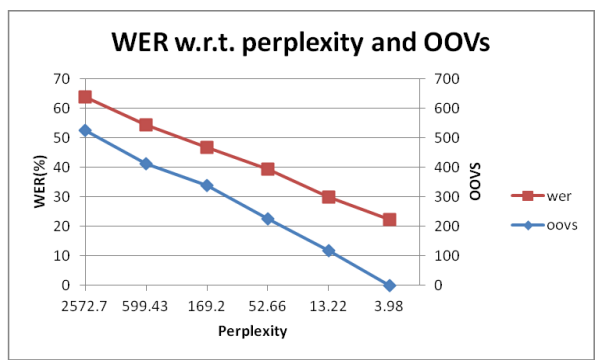

*Figure 1 WER with respect to perplexity and OOVs*

Perplexity typically depends on the training and testing data. However, perplexity does not depend only on the size of training or testing data, but it also depends on the occurrence of test events in the training data. So, the training data is modified by splitting the words in the training data to root-word and suffix or prefix, wherever it is possible. Even though some of the root words and prefix or suffix does not occur together in the training data, these combined words can be generated easily after recognizing the roots, suffixes and prefixes of words.

The language model with Witten-Bell smoothing, which gives less WER without splitting any word, is considered to evaluate the performance of ASR with varying perplexity.

The Language model is generated using Witten-Bell smoothing technique. The perplexity of language model is calculated by varying amount of test text included in the text corpus. The performance of ASR with text corpus for different perplexity is given in the **Error! Reference source not found.**. Figure 1 shows the performance of ASR with respect to perplexity and OOVs obtained without pre-processing text corpus.

TABLE 3 WER FOR THE WITTEN-BELL SMOOTHING LM WITH DIFFERENT PERPLEXITY AND OOVS TOTAL DATA CONSIDERED FOR TESTING IS 2:30 MIN AND WORDS ARE 6814 TOTAL UNIQUE WORDS ARE 3163 AND HMM AS ACOUSTIC MODEL

| S.No | Perplexity | OOVs | WER % | Errors | | | 3-grams hit | 2-grams hit | 1-grams hit |
|---|---|---|---|---|---|---|---|---|---|
| | | | | Insertions | Deletions | Substitutions | | | |
| 1 | 1610.23 | 662 (9.72%) | 61.7 | 167 | 743 | 3265 | 572 (9.30%) | 1919 (31.1%) | 3661 (59.5%) |
| 2 | 2572.70 | 526 (7.72%) | 63.8 | 114 | 997 | 3240 | 617 (9.81%) | 2144 (34.1%) | 3527 (56.0%) |
| 3 | 599.43 | 413 (6.06%) | 54.36 | 106 | 865 | 2733 | 1888 (29.5%) | 1749 (27.32%) | 2764 (43.18%) |
| 4 | 169.20 | 338 (4.96%) | 46.84 | 100 | 756 | 2336 | 3096 (47.81) | 1290 (19.21%) | 2090 (32.97%) |
| 5 | 52.66 | 226 (3.32%) | 39.30 | 86 | 668 | 1924 | 4220 (64.0%) | 900 (13.6 %) | 1468 (22.28%) |
| 6 | 13.22 | 117 (1.72%) | 29.84 | 81 | 551 | 1401 | 5593 (83.%) | 404 (6.03%) | 700 (10.45%) |
| 7 | 3.98 | 0 (0.00%) | 22.17 | 73 | 437 | 1001 | 6812 (99.9%) | 1 (0.01%) | 1 (0.01%) |

IV. PERFORMANCE OF ASR ACCURACY WITH PREPROCESSING THE TEXT CORPUS WITH THE SUPERVISED AND UNSUPERVISED STEMMING TECHNIQUES

*A. Supervised Stemming Techniques*

For Indian languages, particularly for Telugu, most of the base/root words are inflected to match with the context of tense, plural or singular, and gender. The base words are either nouns or verbs. The nouns can be modified in two ways: First, the singular nouns are inflected to make plural. Second, morphological modification through case markers: nominative, genitive, dative, accusative, vocative, instrumental and locative[30]. The morphology of Telugu words as described in literature[31][32]. For example, particular type of suffix will be added to the verbs (base word) ending with "uu" sound to make new words, depending on the gender, person, and number as shown in the Table *4*.

If we have the base words along with their possible inflections (suffixes and prefixes) mentioned in text corpus, the generated LM will not have the OOVs related to base words and their inflections. If we include, all possible inflected words of root/base, which are not available in the training text corpus into the lexicon, the probability of these words is minimal or zero The recognition probability of such set of words is minimal. In Telugu writing system, the minimal unit is orthographic syllable or akshara. It is possible to separate the words and their inflections. Therefore, instead of including every possible inflecting form separately, the base words, prefix and suffix parts are included as individual words in the phonetic lexicon. Very few Telugu words, mostly Sanskrit based ones have prefixes (e.g. aadaraNa vs. anaadaraNa in which 'an' is the prefix)[33]. In words like cittaciwara, mottamodalu the first parts are not prefixes for these words are compound words. Also, in one of the examples listed in the Table *4*, caduwucunnaadu 'he is reading', there are many morphemes and linguists would separate this word as: caduwu-tuu-un-naa-Du to indicate the base, present prog. Tense, male gender [34]. since our focus is on obtaining better accuracy for ASR, by reducing the number of OOVs, we do not need to isolate all the morphemes in this manner. The word 'suffix' and prefix do not refer to all the morpho-phonemic alternations. Language model will assign probability for base words and all the inflecting prefix and suffix parts as words. Then the Language model will also give the probabilities for unseen inflected words as bigrams in the training corpus using smoothing techniques.

Table 4 VERB ENDING WITH UU ( IN GOOGLE TRANSLITERATION IT IS VU) AND ITS SUFFIX BASED ON GENDER AGREEMENT

| S.No | Combined word | Verb ending with uu | Suffix |
|---|---|---|---|
| 1. | చదువు చున్నాడు (Caduvu cunnāḍu) | చదువు(caduvu) | చున్నాడు(Cunnāḍu) |
| 2. | చదువు చున్నది (Caduvu cunnadi) | చదువు(caduvu) | చున్నది(cunnadi) |
| 3. | చదువు చున్నావు (caduvu cunnāvu) | చదువు (caduvu) | చున్నావు (cunnāvu) |
| 4. | చదువుచున్నాను(caduvu cunnānu) | చదువు(caduvu) | చున్నాను(Cunnānu) |
| 5. | చదువు చున్నారు (caduvu cunnāru) | చదువు (caduvu) | చున్నారు (cunnāru) |
| 6. | చదువు చున్నవి (caduvu cunnavi) | చదువు (caduvu) | చున్నవి (cunnavi) |
| 7. | చదువు చున్నాము (caduvu cunnāmu) | చదువు (caduvu) | చున్నాము (Cunnāmu) |
| 8. | చదువు కున్నాడు (caduvu kunnāḍu) | చదువు (caduvu) | కున్నాడు (kunnāḍu) |
| 9. | చదువు కున్నది (caduvu kunnadi) | చదువు (caduvu) | కున్నది (kunnadi) |
| 10. | చదువు కున్నావు (caduvu kunnāvu) | చదువు (caduvu) | కున్నావు (kunnāvu) |
| 11. | చదువు కున్నాను (caduvu kunnānu) | చదువు (caduvu) | కున్నాను (kunnānu) |
| 12. | చదువు కున్నారు (caduvu kunnāru) | చదువు (caduvu) | కున్నారు (kunnāru) |
| 13. | చదువు కున్నవి (caduvu kunnavi) | చదువు (caduvu) | కున్నవి (kunnavi) |
| 14. | చదువు కున్నాము (caduvu kunnāmu) | చదువు (caduvu) | కున్నాము (kunnāmu) |
| 15. | చదువు తున్నాడు (caduvu tunnāḍu) | చదువు (caduvu) | తున్నాడు (tunnāḍu) |
| 16. | చదువు తున్నది (caduvu tunnadi) | చదువు (caduvu) | తున్నది (tunnadi) |
| 17. | చదువు తున్నావు (caduvu tunnāvu) | చదువు (caduvu) | తున్నావు (tunnāvu) |
| 18. | చదువు తున్నాను (caduvu tunnānu) | చదువు (caduvu) | తున్నాను (tunnānu) |
| 19. | చదువు తున్నారు (Caduvu tunnāru) | చదువు (caduvu) | తున్నారు (tunnāru) |
| 20. | చదువు తున్నవి (caduvu tunnavi) | చదువు (caduvu) | తున్నవి (tunnavi) |
| 21. | చదువు తున్నాము (caduvu tunnāmu) | చదువు (caduvu) | తున్నాము (tunnāmu) |
| 22. | చదువుట (caduvuṭa) | చదువు (caduvu) | ట (ṭa) |
| 23. | చదువుతాను (caduvutānu) | చదువు (caduvu) | తాను (tānu) |
| 24. | చదువుతావు (Caduvutāvu) | చదువు (caduvu) | తావు (tāvu) |
| 25. | చదువుము (caduvumu) | చదువు (caduvu) | ము (mu) |
| 26. | చదువుతాము (caduvutāmu) | చదువు (caduvu) | తాము (tāmu) |
| 27. | చదువుతారు (caduvutāru) | చదువు (caduvu) | తారు (tāru) |

Let us consider the inflected words *chan'dan'gaa*, *mukhyan'gaa*, *akhilapakshhan'gaa*, *kein'draman'trigaa*, *kein'dran'gaa*. These four words have different base-words with a fixed suffix. Similarly, some inflected words will have same base-word but different suffixes. Such as, *akhilapakshhan'to* vs *akhilapakshhan'gaa*; *kein'draman'tri'to* vs *kein'draman'tri'gaa*. If, the words are separated at inflections *gaa* and *to*, the word *kein'dran' to*, which is a combination of prefix and suffix of different words, is not available in the training text corpus, will have good statistics. The following table will give a list of possible inflected words for a single base word.

The text is pre-processed to separate the words considering the two grammar rules of Telugu. First considering the 16 inflections given in the Table 5 second the verbs ending with UU phoneme will follow one of the 21 fixed suffixes. Out of 1,60,271 sentences consisting 19,41,832 words and 2,10,000 unique words in the text corpus, 26,213 unique words are identified in the text corpus for splitting in to root/base and suffix or prefix as given in the Table 6.

TABLE 5 LIST OF POSSIBLE INFLECTIONS FOR A SINGLE BASE WORD

| గా (gā) | తో (tō) | పైన (paina) | ను (nu) | గానే (gānē) | లలోని (lalōni) | తాము (tāmu) | తాను (tānu) |
|---|---|---|---|---|---|---|---|
| లో (lō) | కు (ku) | లోని (lōni) | లోనే (lōnē) | తోనే (tōnē) | తోనూ (tōnū) | తారు (tāru) | తావు (tāvu) |

TABLE 6 THE BASE NOUN AND THEIR INFLECTIONS

| S.No | Original word | Word prefix | Word suffix(inflections) case markers |
|---|---|---|---|
| 1 | ఆంధ్రప్రదేశ్‌గా | ఆంధ్రప్రదేశ్ | గా (gā) |
| 2 | ఆంధ్రప్రదేశ్‌లో | ఆంధ్రప్రదేశ్ | లో (lō) |
| 3 | ఆంధ్రప్రదేశ్‌పైన | ఆంధ్రప్రదేశ్ | పైన (paina) |

| 4 | ఆంధ్రప్రదేశలోని | ఆంధ్రప్రదేశ్ | లోని(lōni) |
| 5 | ఆంధ్రప్రదేశలలోని | ఆంధ్రప్రదేశ్ | లలోని(lalōni) |
| 6 | ఆంధ్రప్రదేశ్‌తోనూ | ఆంధ్రప్రదేశ్ | తోనూ(tōnū) |
| 7 | ఆంధ్రప్రదేశ్‌తో | ఆంధ్రప్రదేశ్ | తో(tō) |
| 8 | ఆంధ్రప్రదేశకు | ఆంధ్రప్రదేశ్ | కు(ku) |
| 9 | ఆంధ్రప్రదేశను | ఆంధ్రప్రదేశ్ | ను(nu) |
| 10 | ఆంధ్రప్రదేశలోనే | ఆంధ్రప్రదేశ్ | లోనే(lōnē) |
| 11 | ఆంధ్రప్రదేశ్‌గానే | ఆంధ్రప్రదేశ్ | గానే(gānē) |
| 12 | ఆంధ్రప్రదేశ్‌తోనే | ఆంధ్రప్రదేశ్ | తోనే(tōnē) |

TABLE 7 WER FOR THE WITTEN-BELL SMOOTHING LM WITH DIFFERENT PERPLEXITY AND OOVs TOTAL DATA CONSIDERED FOR TESTING IS 2:30 MIN AND WORDS ARE 7476.

| S.No | Perplexity | OOVs (%) | WER % | Errors | | | 3-grams hit (%) | 2-grams hit (%) | 1-grams hit(%) |
| --- | --- | --- | --- | --- | --- | --- | --- | --- | --- |
| | | | | Insertions | Deletions | Substitutions | | | |
| 1 | 850.12 | 612 (8.18%) | 59.64 | 176 | 937 | 3349 | 1011 (14.72%) | 2473 (36.00%) | 3386 (49.29%) |
| 2 | 1276 | 491 (6.56%) | 62.20 | 136 | 1136 | 3382 | 1127 (16.12%) | 2687 (38.44%) | 3177 (45.44%) |
| 3 | 373.34 | 386 (5.16%) | 53.80 | 135 | 1016 | 2874 | 2442 (34.41%) | 2172 (30.61%) | 2482 (34.98%) |
| 4 | 131.68 | 315 (4.21%) | 46.83 | 124 | 890 | 2490 | 3679 (51.33%) | 1586 (22.13%) | 1902 (26.54%) |
| 5 | 49.41 | 207 (2.77%) | 40.08 | 97 | 808 | 2094 | 4816 (66.20%) | 1119 (15.38%) | 1340 (18.42%) |
| 6 | 15.31 | 108 (1.44%) | 31.02 | 98 | 688 | 1535 | 6217 (84.31%) | 513 (6.96%) | 644 (8.73%) |
| 7 | 5.43 | 0 (0.00%) | 23.63 | 89 | 558 | 1121 | 7480 (99.97%) | 1 (0.01%) | 1 (0.01%) |

From the **Error! Reference source not found.**Table 7 it observed that, the performance of Witten-Bell smoothing degrades when text corpus is pre-processed with supervised stemming, Kneser-Ney smoothing is considered as an alternative for pre-processed data.

TABLE 8 PERFORMANCE OF ASR WITH LM GENERATED WITH THE TEXT CORPUS OF TRAINING DATA AND WITH/ WITHOUT TEST DATA

| | LM generated with the text corpus of training data and | | | | | | | | | | | | | | | | |
|---|---|---|---|---|---|---|---|---|---|---|---|---|---|---|---|---|---|
| | With test data | | | | | | without test data | | | | | | without test data but including OOVs as words | | | | | |
| | Combined words | | split words(IA) | | split words-unsupervised | | Combined words | | split words(IA) | | split words-unsupervised | | Combined words | | split words(IA) | | split words-unsupervised | |
| Type of smoothing | Witten bell | Kneser-Ney | Witten bell | Kneser-Ney | Witten bell | Kneser-ney | Witten bell | Kneser-ney | Witten bell | Kneser-ney | Witten bell | Kneser-ney | Witten bell | Kneser-ney | Witten bell | Kneser–ney | Witten bell | Kneser-ney |
| WER-HMM | 25.16 | 30.03 | 25.36 | 29.97 | 26.06 | 28.12 | 62.24 | 62.50 | 60.39 | 59.72 | 57.82 | 56.23 | 59.61 | 58.96 | 58.41 | 57.32 | 57.01 | 55.23 |
| WER-SGMM | 15.28 | 18.69 | 15.78 | 17.93 | 15.36 | 16.97 | 52.94 | 53.21 | 50.86 | 49.98 | 47.94 | 43.21 | 52.94 | 49.23 | 47.74 | 44.24 | 46.54 | 42.24 |
| Perplexity | 85.00 | 172.75 | 65.94 | 137.07 | | | 3727.91 | 3577.09 | 1500.06 | | | | 6412.57 | 2654.48 | | | | |

To evaluate the performance of ASR, we considered three cases for language model generation. In the first case, test text is included in the text corpus, for the second case, the test text is removed from the text corpus, in third case only OOVs are included as words in the text corpus. Witten-bell and knser-ney are employed in all three cases. Performance of ASR with Witten-Bell and Kneser-Ney Language Model with supervised stemming of pre-processed text corpus is given in the TABLE 8. The results are averaged each time considering the 5 hours 30 minutes speech for testing from the total speech corpus of 65 hours. Whenever text corresponding to the test data is included in generating the language model, Witten-Bell smoothing performs well when compared to Kneser-Ney smoothing.

*B. Unsupervised Stemming Techniques*

The inflected words are bifurcated into, root word with suffix or prefix by forming the rules drawing on the knowledge of the grammar. However, as a set of suffixes and prefixes are common for most of the inflected words, new words are formed by combining two words with "sandhi" or "samaasamu" in Telugu and many other Indian languages like "antarvedi" "Antarangamu", where "anta" is common prefix for these two words. There are some more words that have common suffix or prefix parts, without any meaning. So, statistical techniques for segmentation of inflected words are developed for Hindi and Telugu languages[35][36]. As the knowledge over grammar of language is not used for segmentation, this type of segmentation is called unsupervised segmentation. However, as the segmentation is done purely through statistics, it may not be assured that all the segmented words are generated only from the inflected words.

Supervised segmentation is essential for NLP, as the broken words are used to understand the meaning of a word or sentence. However, the recognized words in the ASR, are combined to form the single word and automatically gives the meaning. So, an unsupervised segmentation of words is proposed in the literature for ASR for Hindi. The same technique with few changes is implemented for segmentation of text corpus. This approach is most similar to that of [36] which employs statistical analysis to discover stems.

This method uses an unsupervised algorithm that automatically discovers stems and suffixes from a corpus vocabulary. It requires a corpus vocabulary, a stem frequency threshold and a suffix frequency threshold as inputs. First step corresponds to splitting all the words in the vocabulary at every character (according to the Unicode encoding) and then define a bi-partite graph that represents all the prefixes and suffices as its partite sets with edges between vertices corresponding to a valid word in the given vocabulary. Then restrict this graph to the maximal set of prefixes and suffixes such that each prefix has minimum number of different suffixes and each suffix has at least minimum number of different prefixes. This maximal set is found using an algorithm Prune given in the[3], which iteratively removes the prefixes and suffixes corresponding to every vertex, whose degree is below the corresponding threshold.

*okkokka saari nijamngaa bayatxa unna imeij eidaitei umntxumndoo manamu daaniki saripootaamaa nijamngaa* ( original )

*okasaari nijamngaa bayatxa unna eidai too umntxumndoo manamu daaniki saripoo tama nijamn gaa* (split)

*okokasaari nijamngaa bayatxa unnaanani epitoo umntxumndoo manamu daaniki saripootaanu maa nijamngaa* (combined)

TABLE 9 THE NUMBER OF UNIQUE WORDS IN THE TRAINING CORPUS

| S.No | Type of splitting | Number of Unique words |
|---|---|---|
| 1 | Without splitting | 2,37,994 |
| 2 | After splitting the inflections with supervision | 2,01,657 |
| 3 | Unsupervised splitting criteria | 1,96,435 |

TABLE 10 WORD ERROR RATE WITH DIFFERENT LANGUAGE MODELING TRAINING CORPUS WITH SGMM AS ACOUSTIC MODELLING

| S.No | Training corpus included for LM generation | WER (%) |
|---|---|---|
| 1 | Without stemming | 18.69 |
| 2 | After stemming the inflections with supervision | 17.93 |
| 3 | unsupervised stemming | 16.97 |
| 4 | Without splitting + unsupervised stemming | 16.42 |

There is a significant reduction in number of unique words from 2,37,994 to 1,96,435 after un supervised stemming as mentioned in Table 9. From the TABLE 8 it is observed that Knser-Ney smoothing is performing well when the text corpus is pre-processed with supervised technique. We considered Knser-Ney smoothing for further analysis. The performance of ASR before and after unsupervised training is shown in **Error! Reference source not found.**. There is 2% improvement in the ASR performance after unsupervised stemming

It is also observed that because of the possibility of one or two words (which are sometimes critical) being mis recognized in a sentence, despite the fact that the words in the following sentences in the same paragraph, are recognized correctly, the context is not understood and for this reason sometimes the readability of the paragraph is lost. When the language models are built by stemming the words in the corpus, some of the critical words are added automatically to the language models and these words added to the language model are recognized properly for improving the readability.

## V. CONCLUSIONS

Some of the frequently used smoothing techniques were evaluated for Telugu language. The Witten-Bell and Kneser-Ney methods performed better than other techniques. Kneser-Ney smoothing technique performed well when the text corpus is pre-processed using both supervised and unsupervised methods. The ASR word accuracy of the ASR system built with the normal text corpus is 81.01%. The ASR word accuracy was improved by 0.76% when the language models were built with normal text corpus was split using supervised method i.e. grammar/morphological rules. Further the ASR word accuracy was improved by 0.94%, when the language models were built with text corpus split by unsupervised method. The improvement in accuracy is attributed to the reduction of OOVs due to splitting the words in the text corpus. Overall, these techniques improved the readability in the transcribed text.

*Alexandria, Egypt November 17-19, 2006*, 2006, pp. 1–7.

[31] Dr.Divakarla venkatawdhani, *telugu in thirty days*. Hyderabad: Andhrapradesh Sahitya Acadamy, 1976.

[32] B.Purushottam; R Srihari Shastri; D VenkataRama, *Vyakarana padakosamu Sastra nighantuvu*. .

[33] B. Krishnamurti and J. P. L. Gwynn, *A Grammar of Modern Telugu*. Oxford: Oxford University Press, 1985.

[34] B. Krishnamurti, *Telugu Verbal Bases*. Motilal Banarsidass Publishers Pvt. Limited, 2009.

[35] A. P. Siva Kumar, P. Premchand, and A. Govardhan, "TelStem:An Unsupervised Telugu Stemmer with Heuristic Improvements and Normalized Signatures," 2011.

[36] A. K. Pandey and T. J. Siddiqui, "An unsupervised Hindi stemmer with heuristic improvements," in *Proceedings of the second workshop on Analytics for noisy unstructured text data - AND '08*, 2008, pp. 99–105.
15